\documentclass[10pt,twocolumn,letterpaper]{article}
\usepackage{cvpr}
\usepackage{times}
\usepackage{epsfig}
\usepackage{graphicx}
\usepackage{amsmath}
\usepackage{amssymb}
\usepackage{marvosym}
\usepackage{booktabs}
\usepackage[utf8]{inputenc} 
\usepackage[T1]{fontenc}    
\usepackage{url}            
\usepackage{booktabs}       
\usepackage{amsfonts}       
\usepackage{nicefrac}       
\usepackage{microtype}      
\usepackage{xcolor}         
\usepackage{tabularx}
\usepackage{subcaption}
\usepackage{tabulary,multirow,overpic,xcolor}
\definecolor{mygray}{gray}{0.92}
\definecolor{baselinecolor}{gray}{.9}
\newcommand{\baseline}[1]{\cellcolor{baselinecolor}{#1}}
\usepackage{listings}
\usepackage{amsthm}
\usepackage{tabulary}
\usepackage{colortbl}
\usepackage{enumitem}
\usepackage{braket}
\usepackage{array}
\usepackage{float}
\usepackage{enumitem}
\usepackage{booktabs}
\usepackage{algorithm}
\usepackage{listings}
\usepackage{tabu}
\usepackage{pifont} 
\newcommand{\cmark}{\ding{51}}%
\newcommand{\xmark}{\ding{55}}%

\def\x{$\times$}
\newcolumntype{x}[1]{>{\centering\arraybackslash}p{#1pt}}
\newcolumntype{y}[1]{>{\raggedright\arraybackslash}p{#1pt}}
\newcolumntype{z}[1]{>{\raggedleft\arraybackslash}p{#1pt}}
\newlength\savewidth\newcommand\shline{\noalign{\global\savewidth\arrayrulewidth
		\global\arrayrulewidth 1pt}\hline\noalign{\global\arrayrulewidth\savewidth}}
\newcommand{\tablestyle}[2]{\setlength{\tabcolsep}{#1}\renewcommand{\arraystretch}{#2}\centering\footnotesize}

\definecolor{linkcol}{RGB}{233, 4, 141}

\definecolor{xycolor}{RGB}{60, 120, 216}
\definecolor{xycolor}{HTML}{0071bc}

\definecolor{wcolor}{RGB}{103, 78, 167}

\definecolor{dcolor}{RGB}{166, 77,21}

\definecolor{gcolor}{RGB}{204, 102, 153}

\definecolor{tcolor}{RGB}{34,139,34}

\definecolor{iterc}{RGB}{91,196,159}

\definecolor{epochc}{RGB}{96,172,252}

\definecolor{eicolor}{RGB}{153, 51, 102}

\definecolor{citecolor}{HTML}{0071BC}
\definecolor{linkcolor}{HTML}{ED1C24}
\definecolor{cvprblue}{rgb}{0.21,0.49,0.74}
\usepackage[pagebackref,breaklinks,colorlinks,citecolor=cvprblue]{hyperref}
\usepackage{lipsum}
\newcommand\blfootnote[1]{%
  \begingroup
  \renewcommand\thefootnote{}\footnote{#1}%
  \addtocounter{footnote}{-1}%
  \endgroup
}
\usepackage[hang]{footmisc}
\usepackage[capitalize]{cleveref}
\crefname{section}{Sec.}{Secs.}
\Crefname{section}{Section}{Sections}
\Crefname{table}{Table}{Tables}
\crefname{table}{Tab.}{Tabs.}
\def\paperID{5291} 
\def\confName{CVPR}
\def\confYear{2024}
\begin{document}

\title{Asymmetric Masked Distillation for Pre-Training Small Foundation Models}

\author{
Zhiyu Zhao\textsuperscript{1,2}
\quad Bingkun Huang\textsuperscript{1,2}\
\quad Sen Xing\textsuperscript{2} 
\quad Gangshan Wu\textsuperscript{1}
\quad Yu Qiao\textsuperscript{2}
\quad Limin Wang\textsuperscript{1,2,~\Letter} \\[0.2cm]
$^1$ State Key Laboratory for Novel Software Technology, Nanjing University
\quad $^2$ Shanghai AI Lab  \\
}

\maketitle
\begin{abstract}
Self-supervised foundation models have shown great potential in computer vision thanks to the pre-training paradigm of masked autoencoding. Scale is a primary factor influencing the performance of these foundation models. However, these large foundation models often result in high computational cost. This paper focuses on pre-training relatively small vision transformer models that could be efficiently adapted to downstream tasks. Specifically, taking inspiration from knowledge distillation in model compression, we propose a new asymmetric masked distillation~(AMD) framework for pre-training relatively small models with autoencoding. The core of AMD is to devise an asymmetric masking strategy, where the teacher model is enabled to see more context information with a lower masking ratio, while the student model is still equipped with a high masking ratio. We design customized multi-layer feature alignment between the teacher encoder and student encoder to regularize the pre-training of student MAE. To demonstrate the effectiveness and versatility of AMD, we apply it to both ImageMAE and VideoMAE for pre-training relatively small ViT models. AMD achieved 84.6\% classification accuracy on IN1K using the ViT-B model. And AMD achieves 73.3\% classification accuracy using the ViT-B model on the Something-in-Something V2 dataset, a 3.7\% improvement over the original ViT-B model from VideoMAE. We also transfer AMD pre-trained models to downstream tasks and obtain consistent performance improvement over the original masked autoencoding. The code and models are available at \href{https://github.com/MCG-NJU/AMD}{https://github.com/MCG-NJU/AMD}.
\end{abstract}
\blfootnote{\Letter: Corresponding author (lmwang@nju.edu.cn).}
\section{Introduction}
\label{sec:intro}
In recent years, self-supervised learning (SSL)~\cite{chen2020simple, caron2021emerging, chen2021exploring, grill2020bootstrap, he2020momentum,wang2022importance} has witnessed great success and outperformed its supervised counterparts. With the success in masked language modeling (MLM)~\cite{Devlin2019BERTPO}, masked image modeling has become popular in computer vision for self-supervised representation learning. For example, BeiT~\cite{bao2021beit}, SimMIM~\cite{xie2022simmim}, and MAE~\cite{he2021masked} are proposed to image masked pre-training, and MaskFeat~\cite{wei2022masked}, VideoMAE~\cite{videomae}, MAE-ST~\cite{feichtenhofer2022masked}, VideoMAEv2~\cite{wang2023videomaev2}, and MGMAE~\cite{huang2023mgmae} are developed for video masked representation learning. This simple pipeline of masking and reconstruction has shown excellent performance on downstream tasks such as image classification, object detection, semantic segmentation, and action recognition.
\begin{figure}[t]
    \includegraphics[width=1.0\linewidth]{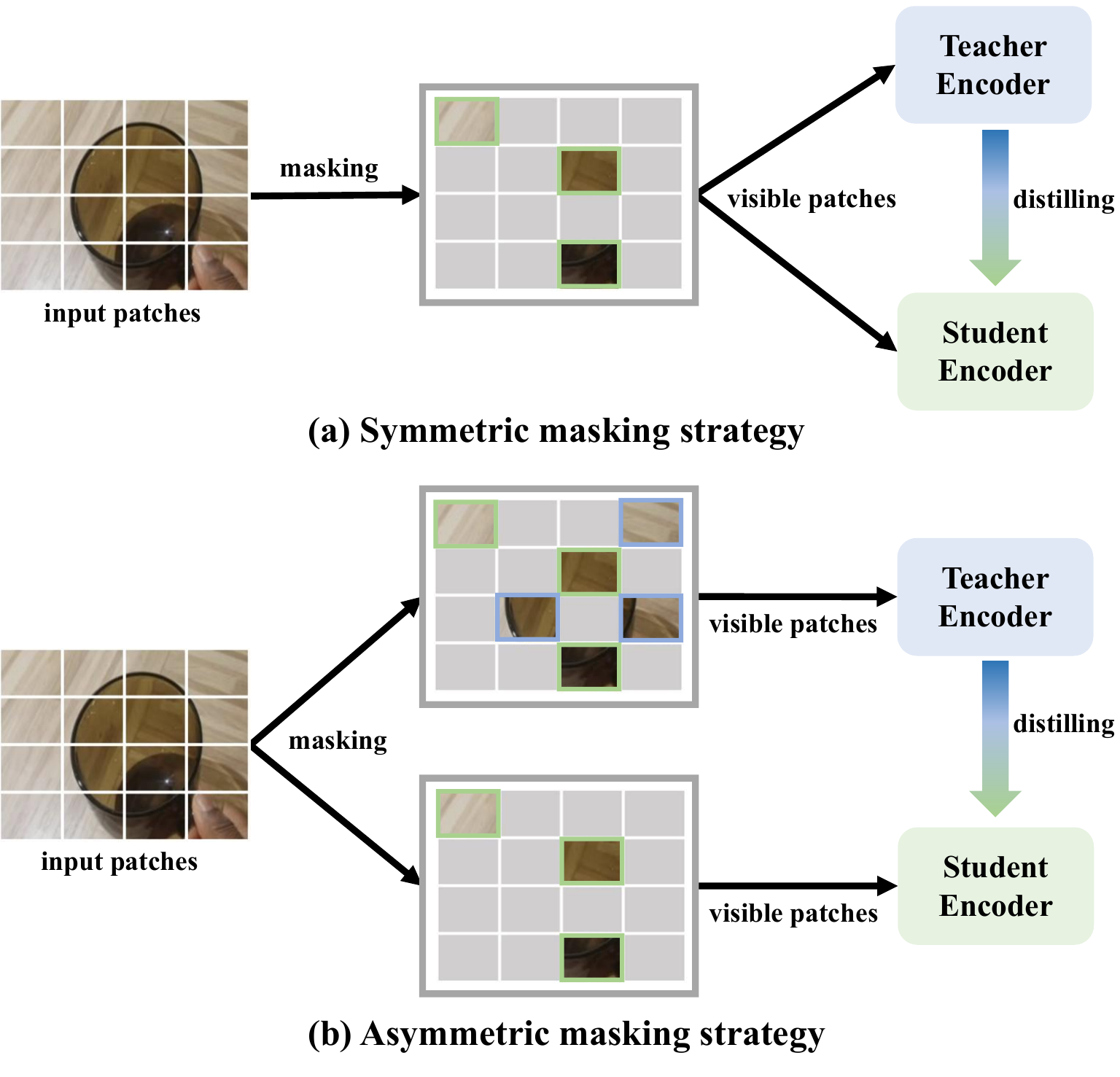}
    \caption{\textbf{Comparison of symmetric and asymmetric masking strategy.} The asymmetric masking strategy allows the teacher to acquire more contextual information than the students.} 
    \label{fig:first}
    \vspace{-2em}
\end{figure}
However, some issues still remain for the masked autoencoding framework. First, the encoder often operates on a small portion of visible tokens with a high masking ratio (\eg, 75\% for image and 90\% for video). This high masking could increase the difficulty of the pre-training tasks and might encourage the encoder to capture more useful high-level information for reconstruction. We argue that this high masking ratio might also lose some important and detailed structure information, leading to the pre-trained model capturing incomplete and biased visual information. Second, the masked autoencoding often requires the ViT backbones of high capacities (\eg, ViT-Large and ViT-Huge) to unleash the power of the masked pre-training. These large ViT models take high computational cost and memory consumption during fine-tuning on the variety of downstream tasks. This high cost is particularly severe for video input as video transformer takes multiple frames as inputs. We think that we should pay more attention to the relatively small ViT models in masked autoencoding, which could have higher efficiency and more application potential in downstream tasks.

To overcome the above issues in masked autoencoding, we resort to the general paradigm for knowledge distillation~\cite{hinton2015distilling} in model compression. It provides an effective {\em teacher} to {\em student} training framework to transfer the dark knowledge in powerful models to the lightweight student model. We extend this idea to the masked autoencoding paradigm to build a more efficient and effective pre-trained model, which could be applied to a variety of downstream tasks. Wei~\etal~\cite{Wei2022ContrastiveLR} applies the feature alignment to distill the unsupervised pre-trained model, but they found that this method yielded little benefit for the MAE pre-trained model, as it already had diverse attention heads. Recently, Bai~\etal~\cite{bai2022masked} proposed a scheme for MAE distillation~(DMAE) where feature distillation is performed alongside pre-training for the reconstruction task. DMAE allows the student and the teacher to receive the symmetric unmasked patches so that features can be aligned directly and the computational complexity of the teacher model can be reduced. But the same masking for both teacher and student limits the teacher from gathering more context information from inputs, and still faces the information loss risk as mentioned above. 

Based on the above analysis, we propose an asymmetric masked distillation structure for MAE pre-training and the goal is to obtain a small but robust pre-trained MAE model. The masking ratio of the student remains at its default setting, while the masking ratio of the teacher is relatively reduced. And the unmasked patches of the student are a subset of those of the teacher, as in Figure \ref{fig:first}. This asymmetric masked distillation maintains the difficulty of the reconstruction task for the students during the pre-training process and also allows the teacher to receive more context information that can be transferred to the student. However, it is not reasonable to significantly reduce the masking ratio of the teacher, as the teacher takes up a high amount of computational resources. Hence we have made a proper compromise between the masking ratio of the teacher and the computational cost. In our asymmetric structure, the visible patches are divided into two types, one is visible to both student and teacher, and the other is visible only to the teacher. To align these two types of features, a serial alignment strategy is applied.

With this efficient design of asymmetric masked distillation, we were able to achieve a performance of 73.3\% on SSV2 based on VideoMAE, leaving a gap of only 1\% with the larger teacher. And we achieved a performance of 84.6\% on IN1K based on ImageMAE. In summary, our main contribution is as follows:
\begin{itemize}[]
    \item We proposed an asymmetric masked distillation strategy for MAE pre-training, which allows the teacher to acquire more context information while maintaining the reconstruction difficulty of the student MAE model.
    \item We presented a serial feature alignment manner for the asymmetric masking strategy to achieve sufficient knowledge distillation for MAE pre-training.
    \item We have successfully employed asymmetric masked distillation to obtain the small and robust AMD with improved distillation efficiency and improved transfer performance on downstream tasks.
\end{itemize}
\section{Related work}
\label{sec:related}
\par{\noindent \textbf{Vision masked modeling.}} 
Recovering an original image from a broken image has recently been introduced as an efficient pre-training paradigm. Early work applied a similar approach for image denoising~\cite{vincent2008extracting,vincent2010stacked} or image inpainting~\cite{guillemot2013image}. Along with the success of Transformer~\cite{vaswani2017attention} in computer vision, recent works~\cite{chen2020generative,bao2021beit,dong2023peco,wei2022masked,He2022MaskedAA,xie2022simmim} have attempted to apply the Vision Transformer~(ViT~\cite{dosovitskiy2020image}) to the masked autoencoder. With the success of BERT~\cite{Devlin2019BERTPO} in proposing a generative task as a pre-training target based on the transformer in NLP, BEiT~\cite{bao2021beit} proposes the masked image modeling based on ViT,  which treats image patches as words. The reconstruction target can be divided into high-level features and low-level features. In terms of high-level reconstruction, BEiT performs a two-stage process as the reconstruction target is the discrete token which requires a pre-trained tokenizer~\cite{ramesh2021zero}. Similarly, PeCO~\cite{dong2023peco} has improved the VAE pre-training by encouraging perceptual similarity.

In terms of low-level reconstruction, Maskfeat~\cite{wei2022masked} proposed the one-stage pre-training approach with a reconstruction target of histograms of oriented gradients~(HOG). The reconstruction target of MAE~\cite{He2022MaskedAA} is pixels, and this remarkable work proposes an asymmetric autoencoder structure that leverages a high masking rate to reduce the computational overhead and make scalability possible. SimMIM~\cite{xie2022simmim} extends MIM into the Swin Transformer~\cite{liu2021swin} at the cost of a heavier encoder, but which utilises a simple projection head to predict pixels.

Recent works~\cite{videomae,feichtenhofer2022masked,wang2023videomaev2,huang2023mgmae} have extended MAE from image to video due to its excellent scalability. VideoMAE~\cite{videomae} applies the tube masking strategy to the video data and adopts joint spatio-temporal attention in the Transformer block to extract video features, which performs excellently on the Something-in-Something V2~\cite{Goyal2017TheS} dataset. Our work follows VideoMAE and proposes an asymmetric distillation scheme for the pre-trained model.
\par{\noindent \textbf{Knowledge distillation and self distillation.}} 
Knowledge distillation~(KD) is an efficient way to compress models and was first proposed by Hinton~\etal in ~\cite{hinton2015distilling}. A common distillation technique is to utilise the logit output from the teacher as a medium for transferring knowledge~\cite{hinton2015distilling,Ba2014DoDN,Chen2017LearningEO,yang2022vitkd,Touvron2021TrainingDI,Purwanto2019ExtremeLR,Wu2022TinyViTFP}. And a temperature factor was introduced to align the soft labels with Kullback-Leibler divergence loss. However, the logit-based distillation approach can only be applied to fine-tuned models, which would damage the generalization ability of unsupervised pre-train models.

In addition to logit, researchers have also exploited intermediate features of the model for feature distillation~\cite{Romero2015FitNetsHF,Heo2019ACO,Wei2022ContrastiveLR,yang2022vitkd,Yang2022FocalAG,bai2022masked,Hou2022MILANMI,peng2022maskdistill,liu2022dbot}. ViTKD~\cite{yang2022vitkd} used logit alignment in conjunction with feature alignment and performed well on the fine-tuned model using supervised information. Wei~\etal~\cite{Wei2022ContrastiveLR} have successfully applied feature alignment to the contrastive-based self-supervised learning methods. By analysing the optimization friendliness properties, they conclude that MAE can hardly benefit through the direct feature distillation.
dBOT~\cite{liu2022dbot} proposed a multi-stage distillation method and use a randomly initialized model as the teacher model. MaskDistill~\cite{peng2022maskdistill} reconstructed the normalized semantic features of the teacher.
DMAE~\cite{bai2022masked} adopted a symmetric masking approach based on MAE for feature alignment during student pre-training with pixel reconstruction task.
G2SD~\cite{huang2023generic} aligns the features of the decoder and applies two-stage distillation to best exploit the teacher's knowledge.
MVD~\cite{wang2022masked} simultaneously employs both image model and video model to distill the MAE model.

A similar style of distillation has emerged in the recent MAE self-distillation works~\cite{Chen2022sdae,chen2023context,el2021large,zhou2021ibot}. They constructed the student and teacher models in a two-stream structure, typically designed with the student and teacher masked patch in a complementary relationship. SdAE~\cite{Chen2022sdae} has successfully accelerated the self-distillation procedure of MAE by analysing information bottleneck and applying the multi-fold masking strategy for the teacher branch. Inspired by those works, our work applies an asymmetric mask strategy to perform feature-based distillation during the pre-training phase of MAE.

\section{Method}
In this section, we first review the structure of VideoMAE~\cite{videomae}, then introduce our AMD framework and introduce the asymmetric mask approach, and finally describe how we perform feature alignment to accomplish the sufficient knowledge distillation.

\subsection{Revisiting VideoMAE}
VideoMAE~\cite{videomae} extends the masked autoencoder from the image domain to the video domain. Formally, each input video will be randomly sampled into a clip with $T$ frames $V\in\mathbb{R}^{T\times H \times W\times3}$. The sampling stride $\tau$ is set up specifically for the dataset.

\par{\noindent \textbf{Patch embedding.}}
Due to the extra time dimension of the video data, VideoMAE treats a $2\times16\times16\times3$ cube as a patch namely joint space-time cube embedding~\cite{arnab2021vivit}. Then the 3D-CNN is employed to process the patches, performing convolution without overlap, to obtain a total of $\hat{T}\times\hat{H}\times\hat{W}$ tokens, whose dimension is mapped to $D$, where $\hat{T}=\frac{T}{2}, \hat{H}=\frac{H}{16}, \hat{W}=\frac{W}{16}$. This allows tokens to be handled in a sequential perspective, whose length is $N$. 

\par{\noindent \textbf{Masking strategy.}}
Due to the redundancy of information in the video data, a higher masking ratio $r$~(\eg 90\%) is applied by VideoMAE. To further reduce information leakage, VideoMAE employs tube masking to mask multiple frames. Specifically, a random binary mask map $\Tilde{M}\in\mathbb{R}^{\hat{H}\times \hat{W}}$ is first generated in token units. To ensure that a given token in the spatial dimension is masked in all temporal dimensions, VideoMAE simply repeats $\Tilde{M}$ in the temporal dimension $\hat{T}$ times to obtain the final mask map $M\in\mathbb{R}^{\hat{T}\times\hat{H}\times \hat{W}}$. We then flatten $M$ into a binary one-dimensional sequence $\hat{M}\in\mathbb{R}^{N\times 1}$ where $1$ means that the token needs to be masked and $0$ means that it is visible and let $P^{\textit{\textit{vis}}}$ denotes the unmasked token indexes.
\par{\noindent \textbf{Encoder: feature extractor.}}
The encoder is a vanilla ViT that takes the sequence of visible tokens after adding the fixed 1D position encoding~\cite{vaswani2017attention}. It is notable that VideoMAE makes no use of the \texttt{[CLS]} token~\cite{Devlin2019BERTPO}. To allow any two tokens in the entire input sequence to interact with each other, VideoMAE applies the joint space-time attention mechanism~\cite{Liu2022VideoST}. The latent features extracted by the encoder are denoted as $F=\{f^i\in\mathbb{R}^{D}\}_{i=1}^{\hat{N}}$. When applied to downstream tasks, the encoder acts as a feature extractor.

\par{\noindent \textbf{Decoder: Pixel reconstructor.}}
The task of the decoder is to reconstruct the input, which requires the masked patches to be restored in the form of pixels. The depth of the decoder is usually shallower and less wide than the encoder. A linear layer is used to map the dimension of the latent features $F$ to the width of the decoder $D^{\textit{dec}}$. The latent features are then concatenated with the learnable \texttt{[MASK]} tokens under the guidance of position and the fixed 1D position encoding is added to them. The decoder is also a vanilla ViT with joint space-time attention, and the output would go through a projection layer to align the dimensions to the original video, which is formed as $\hat{V}\in\mathbb{R}^{T\times H \times W\times3}$.

\begin{figure*}[t]
  \centering
   \includegraphics[width=1.0\linewidth]{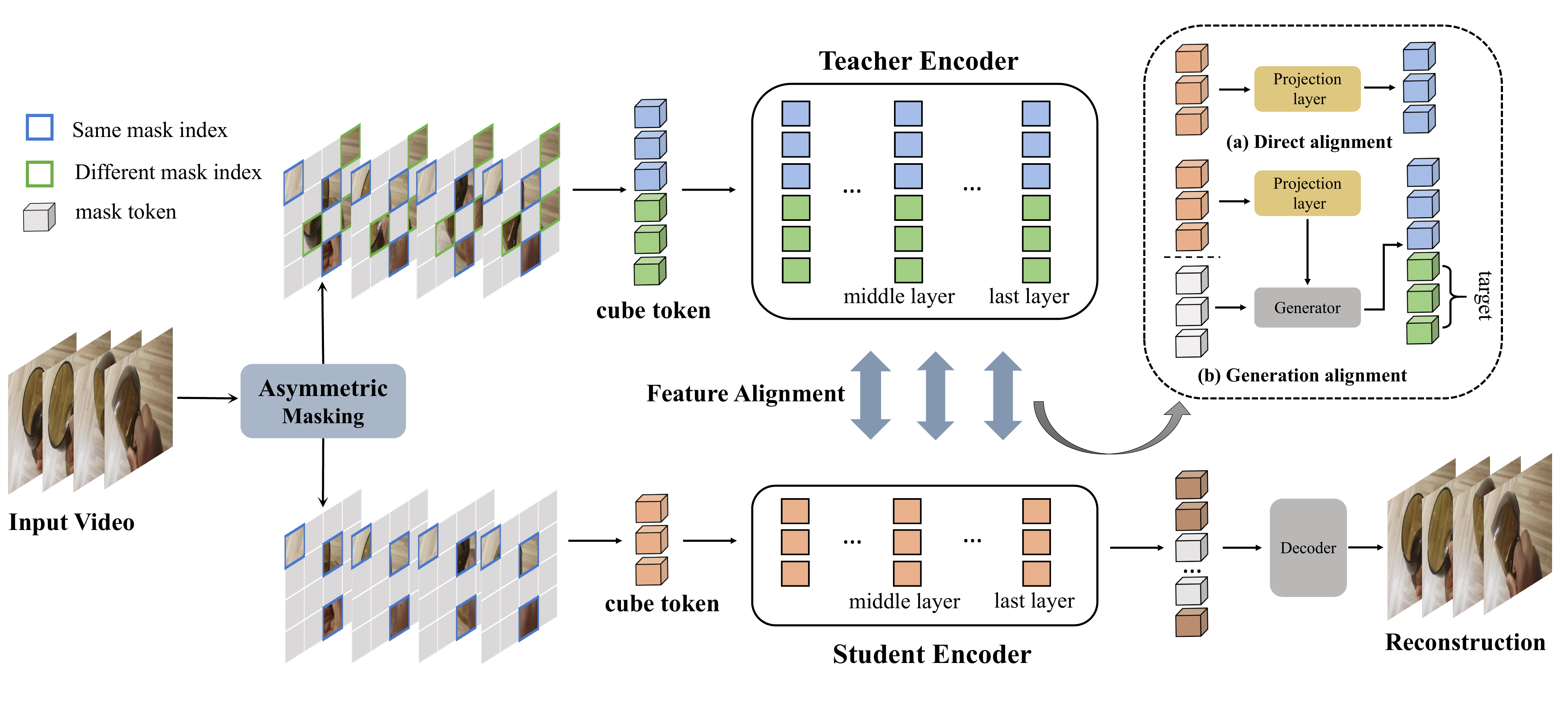}
   \caption{{\bf Pipeline of Asymmetric Masked Distillation~(AMD)}. We present an asymmetric masking strategy to transfer the knowledge of teacher pre-trained models to the student masked pre-training. Our asymmetric masking strategy allows a lower masking ratio for the teacher to enable extracting richer visual information. The richer visual information could be used as guidance information to regularize the student masked pre-training and results in a more powerful pre-trained model, that could benefit a variety of downstream tasks.}
   \label{fig:overview}
\end{figure*}

\par{\noindent \textbf{Objective function.}}
Following MAE, the pre-training task of VideoMAE is to reconstruct pixels. The loss function applied to the reconstruction is mean square error~(MSE) loss, and the reconstruction target is normalised in the token level. The objective function is denoted as:
\begin{equation}
L_{recon}=\frac{1}{rN}\sum_{p\in \Bar{P}^{\textit{vis}}}\left|\text{norm}(V(p))-\hat{V}(p)\right|^2,
\end{equation}
where $\Bar{P}^{\textit{vis}}$ denotes the masked token indexes.

\subsection{Overview of AMD}
Building on the VideoMAE, we propose an asymmetric masked distillation for the MAE pre-training. 
\par{\noindent \textbf{Overview.}} The architecture overview is shown in Figure \ref{fig:overview}. The overall framework of AMD is a two-stream distillation structure with a student branch and a teacher branch, and the teacher is a larger MAE pre-trained model. It is worth noting that AMD only works with the encoder of the teacher for distillation, whereas the students were required to accomplish the pixel reconstruction task with a decoder. Feature distillation occurs between the corresponding layers of the student and teacher models, and AMD employs both direct alignment and generation alignment in a serial way to respond to our asymmetric masking strategy.
\par{\noindent \textbf{Asymmetric mask.}}
A downsampled video clip $V\in\mathbb{R}^{T\times H \times W\times3}$ is the input of our AMD. After performing the cubic patch embedding, our asymmetric mask strategy is then applied to $V$, resulting in the input sequence of the student and the teacher, respectively. The length of the input sequence of the student is shorter than that of the teacher, and the visible token index of the student $P^{\textit{vis}}_{\textit{stu}}$ is a subset of that of the teacher $P^{\textit{vis}}_{\textit{tea}}$. The details are described in \cref{ama}.
\par{\noindent \textbf{Serial feature alignment.}}  It is a feature alignment strategy designed for the asymmetric mask, which combines direct alignment and generation alignment with a shared projection function. The two types of alignment are described in \cref{fd}. Specifically, for a particular layer $l$ of the student, the corresponding layer of the teacher is $l^*$. We use $z$ to represent the features extracted from ViT the feature can be extracted as $z^l_{\textit{stu}}$, $z^{l^*}_{\textit{tea}}$ respectively.

The first step of direct alignment is to apply the projection function $\phi(\cdot)$ to the student to align the dimension of the teacher. In practice, we employ a linear layer for this purpose. Thus we obtained the student features $\Tilde{z}^l_{\textit{stu}} = \phi\left(z^l_{\textit{stu}}\right)$ used for direct alignment.
We then serially use these features to continue with the generation feature alignment $\mathcal{G}(\Tilde{z}^l_{\textit{stu}})$. This serial alignment can appropriately reduce the difficulty of the generation alignment task.
\subsection{Asymmetric Masked Architecture}
\label{ama}

The asymmetry lies in the difference between the inputs of the student and the teacher. Specifically, the masking ratio of the teacher $r_{\textit{tea}}$ is lower than the masking ratio of the student $r_{\textit{stu}}$. Firstly, we generate the token-wise mask map for the student and teacher, and then get the visible token indexes of the student $P^{\textit{vis}}_{\textit{stu}}=\{p_{\textit{stu}}^i\}_{i=1}^{\hat{N}_{\textit{stu}}}$ where $\hat{N}_{\textit{stu}}=(1-r_{\textit{stu}})N$ and the teacher $P^{\textit{vis}}_{\textit{tea}}=\{p_{\textit{tea}}^i\}_{i=1}^{\hat{N}_{\textit{tea}}}$ where $\hat{N}_{\textit{tea}}=(1-r_{\textit{tea}})N$.
The relationship between the two of them is formalized as:
\begin{equation}
    P^{\textit{vis}}_{\textit{stu}}\subsetneqq P^{\textit{vis}}_{\textit{tea}}.
\end{equation}
Thus the teacher can acquire more context information while preserving exactly what the student can receive.

\subsection{Feature Distillation}
\label{fd}
Our work focuses on the pre-training of MAE and employs a serial feature alignment to handle asymmetric masked distillation. Specifically, direct feature alignment and generated feature alignment are performed sequentially. Each of the two types of feature alignment manner is described below.
\par{\noindent\textbf{Direct alignment.}}
Assume that under the condition of symmetric masking strategy, which means that the input lengths $\hat{N}$ of the teacher and student are the same and they share the same set of visible token indexes $P^{\textit{vis}}$. The features of the student and the teacher obtained from a pair of corresponding layers are respectively denoted as $z^l_{\textit{stu}}\in\mathbb{R}^{\hat{N}\times D_{\textit{stu}}},z^{l^*}_{\textit{tea}}\in\mathbb{R}^{\hat{N}\times D_{\textit{tea}}}$, where $D$ means the dimension of features, $l$ indicates the layer index of the student model and $l^*$ denotes the layer index corresponding to the student layer~(\eg the middle or last layer).
The feature dimension of the teacher is generally larger, so a projection function $\phi_d$ for alignment is necessary. The loss function for a single-layer direct alignment can be defined as:
\begin{equation}
    L_{\textit{dir}}=\frac{1}{\hat{N}}\sum_{p_i\in P^{\textit{vis}}}\left|z^{l^*}_{\textit{tea}}(p_i)-\phi_d\left(z^{l}_{\textit{stu}}(p_i)\right)\right|^2,
\end{equation}
where $z(p)$ indicates that the feature is extracted from the $p$-th token of the input sequence.
\par{\noindent \textbf{Generation alignment.}}
The generation alignment can be applied when the input length of the student and teacher do not match. The difference in length between the two is denoted as $\hat{N}_{\textit{diff}}=\hat{N}_{\textit{tea}}-\hat{N}_{\textit{stu}}$. And in terms of the visible token index, we denote $P^{\textit{diff}}=P^\textit{vis}_{\textit{tea}}\setminus P^\textit{vis}_{\textit{stu}}$.
We denote the features of the student and the teacher obtained from a pair of corresponding layers by $z^l_{\textit{stu}}$, $z^{l^*}_{\textit{tea}}$ respectively.
As the feature dimensions of the teacher and the student are not aligned, a simple linear layer $\phi_g(\cdot)$ was employed to map the dimension of the student features.

Since the teacher has more tokens than the student, our work utilises the student model to generate tokens to align with the teacher. Specifically, our generator $\mathcal{G}(\cdot)$ is a decoder-like structure with multi-head self-attention~(MHA). Similar to the decoder, we need to concatenate the input with the $\texttt{[MASK]}$ token $z_m$ and add the fixed 1D position encoding~(PE). The generation process can be described as:
\begin{equation}
    \Tilde{z}^l_{\textit{stu}}=\phi_g\left(z^l_{\textit{stu}}\right)\in\mathbb{R}^{\hat{N}_{\textit{stu}}\times D_{\textit{tea}}},
\end{equation}
\vspace{-2em} 
\begin{equation}
\mathcal{G}\left(\Tilde{z}^l_{\textit{stu}}\right)=\text{MHA}\left(\text{concat}\left(\Tilde{z}^l_{\textit{stu}},\text{repeat}\left(z_m\right)\right)+\text{PE}\right).
\end{equation}
It is worth noting that in a normal decoder structure, the repeat number of the $\texttt{[MASK]}$ token is the number of all invisible tokens. However, in our work, we only repeat the $\texttt{[MASK]}$ token $\hat{N}_{\textit{diff}}$ times in order to reduce the generation of redundant features during alignment.
We employ the MSE as the loss function for the training of feature alignment for a single layer:
\begin{equation}
    L_{\textit{gen}}=\frac{1}{\hat{N}_{\textit{diff}}}\sum_{p_i\in P^{\textit{diff}}}\left|z^{l^*}_{\textit{tea}}(p_i)- \mathcal{G}\left(\Tilde{z}^l_{\textit{stu}}\right)(p_i)\right|^2,
\end{equation}
\subsection{Objective function}
\par{\noindent \textbf{Multi-layer alignment.}} We can choose more than one layer for feature alignment, in practice, we have experimented with two layers, the middle layer and the last layer. As the distribution of features within different layers is varied, the parameters used for alignment are not shared between the different layers. Therefore, when aligning multiple layers, the loss function for direct alignment is rewritten as:
\begin{equation}
     L_{\textit{dir}}=\sum_l\frac{1}{\hat{N}_{\textit{stu}}}\sum_{p_i\in P^{\textit{vis}}_\textit{stu}}\left|z^{l^*}_{\textit{tea}}(p_i)-\phi_l\left(z^l_{\textit{stu}}(p_i)\right)\right|^2,
\end{equation}
where $\phi_l$ represents the projection function for layer $l$. The loss function for generation alignment is rewritten as:
\begin{equation}
    L_{\textit{gen}}=\sum_l\frac{1}{\hat{N}_{\textit{diff}}}\sum_{p_i\in P^{\textit{diff}}}\left|z^{l^*}_{\textit{tea}}(p_i)- \mathcal{G}_l\left(\Tilde{z}^l_{\textit{stu}}\right)(p_i)\right|^2,
\end{equation}
where $\mathcal{G}_l$ represents the generator for layer $l$.
\par{\noindent \textbf{Overall.}} Students are required to reconstruct pixels while completing serial feature alignment, so the overall loss function for AMD is defined as:
\begin{equation}
    L_{\textit{total}}=L_{\textit{recon}}+L_{\textit{dir}}+L_{\textit{gen}}.
\end{equation}
\section{Experiments}

\subsection{Datasets}
Following VideoMAE, we evaluate our AMD on five video datasets: Something-Something V2~(SSV2)~\cite{Goyal2017TheS}, Kinetics-400~(K400)~\cite{carreira2017quo}, UCF101~\cite{soomro2012ucf101}, HMDB51~\cite{kuehne2011hmdb} and AVA~\cite{gu2018ava}. SSV2 contains around 169k training videos and 20k validation videos belonging to 174 action classes. K400 contains about 240k training videos and 20k validation videos from 400 categories. UCF101 and HMDB51 are two small datasets which contain around 9.5k/3.5k training videos and 3.5k/1.5k validation videos respectively. AVA contains 211k training videos and 57k validation videos which is a benchmark for the spatio-temporal localization task. AMD is only pre-trained on SSV2 and K400 and other datasets are used for fine-tuning only. The implementation details are presented in the appendix.
\subsection{Ablation Study}
\label{abla}
In this section, we perform ablation experiments on AMD with 16-frames vanilla ViT-B, and all results are obtained on SSV2. We run 200 epochs per experiment. The ViT-L model of VideoMAE with 2400 epochs pre-trained on SSV2 is employed as the teacher model for distillation. 

Since our distillation strategy is built into the MAE pre-training process, we fixed the masking ratio of the student at 90\% which is the default setting of VideoMAE, in order not to damage the reconstruction difficulty.
As for the fine-tuning setting, the sampling strategy adopted for SSV2 is uniform sampling~\cite{tsn_journal} and a $2$ clips \x $3$ crops test is used to obtain the final results.

\par{\noindent \textbf{Generator depth.}} 
The generator is a decoder-like structure and we compared the effect of different generator depths on the performance of the model in Table~\ref{tab:decoder_depth}. A deeper generator provides better classification performance but also brings more computational overhead. The performance gain from 1 layer to 2 layers is significant, but the gain from 2 layers to 4 layers is lower and more time-consuming. We have chosen a depth of 2 as the default setting, which compromises performance and computational complexity.
\begin{table}[t]
\vspace{-.2em}
\subfloat[
\textbf{Generator depth}. Our default 
choice is a reasonable compromise between performance and computational overhead.
\label{tab:decoder_depth}
]{
\hspace{-0.5em}
\begin{minipage}{0.42\linewidth}{\begin{center}
\tablestyle{1pt}{1.1}
\begin{tabular}{x{30}x{28}x{28}}
\toprule
\textbf{Block} & \textbf{Top-1} & \textbf{Time} \\
\midrule
1 &  72.06 & 12.92h\\ 
\baseline{2} & \baseline{72.38} & \baseline{13.33h}\\
4 & \textbf{72.45} & 15.33h\\
\bottomrule
\end{tabular}
\end{center}}
\end{minipage}
}
\hspace{1em}
\subfloat[
\textbf{Layer for alignment}. We compare different layer numbers for alignment. Our choice of alignment layers is the middle and the last layer.
\label{tab:layers}
]{
\begin{minipage}{0.50\linewidth}{
\begin{center}
\tablestyle{2pt}{1.1}
\begin{tabular}{x{24}x{24}x{24}x{24}}
\toprule
\textbf{Layers} & \textbf{Stu.} & \textbf{Tea.} & \textbf{Top-1} \\
\midrule
1 & 12 & 24 & 71.93 \\
 \baseline{}& \baseline{6} &  \baseline{12} &\baseline{}  \\
\multirow{-2}*{\baseline{2}}&\baseline{12} & \baseline{24} &\multirow{-2}*{\baseline{\textbf{72.38}}} \\
\bottomrule
\end{tabular}
\end{center}
}\end{minipage}
}
\\
\subfloat[
\textbf{Masking ratio}. Student's ratio is fixed at 90\%. Our default choice trade-offs performance and computational overhead.
\label{tab:mask_ratio}
]{
\hspace{-0.5 em}
\begin{minipage}{0.42\linewidth}{\begin{center}
\tablestyle{2pt}{1.1}
\begin{tabular}{x{35}x{20}x{24}}
\toprule
\textbf{Tea. ratio} & \textbf{Top-1} & \textbf{Time} \\
\midrule
85\% & 72.17 & 12.08h \\
\baseline{75\%} & \baseline{72.38} & \baseline{13.33h} \\
60\% & 72.44 & 15.47h\\
45\% & \textbf{72.48} & 18.93h\\
\bottomrule
\end{tabular}
\end{center}}
\end{minipage}
}
\hspace{1em}
\subfloat[
\textbf{Distill manner.} AMD works best when using a serial approach combining direct alignment and generative alignment for feature alignment.
\label{tab:method}
]{
\hspace{-0.5em}
\begin{minipage}{0.50\linewidth}{
\begin{center}
\tablestyle{1pt}{1.1}
\begin{tabular}{y{70}x{21}x{22}}
\toprule
\textbf{Method} & \textbf{Top-1} & \textbf{Time} \\
\midrule
\textbf{D}irect only  & 71.97 & 10.83h\\
\textbf{G}eneration only & 71.94 &  12.92h\\
\textbf{D+G}~(in parallel way) & 72.25 & 13.33h \\
\baseline{\textbf{D+G}~(in serial way)} & \baseline{\textbf{72.38}} &\baseline{13.33h}  \\
\bottomrule
\end{tabular}
\end{center}}
\end{minipage}
}
\vspace{0em}
\caption{\textbf{Ablation experiments on Something-Something V2.} All models are trained and timed for 200 epochs on 16 A100 GPUs. The student model is the vanilla ViT-B and the teacher model is the ViT-L of videoMAE\cite{videomae} with 2400 epochs pre-trained on SSV2. The default choice for our AMD is colored in \colorbox{baselinecolor}{gray}.}
\label{tab:ablations}
\vspace{-1em}
\end{table}
\par{\noindent \textbf{Alignment layer selection.}}
When performing feature alignment, we only consider the middle and last layers, which for ViT-B are the 6-th and 12-th layer. We have experimented with different alignment layers in Table~\ref{tab:layers}. When training 200 epochs, we found that aligning two layers simultaneously could produce higher distillation benefits than aligning only one layer. The two are similar in terms of time spent.
\par{\noindent \textbf{Masking ratio.}} In knowledge distillation, we expect the teacher to transfer more context information to the student, which can be achieved by adjusting the masking ratio of the teacher. The comparison is shown in Table~\ref{tab:mask_ratio}. A lower masking ratio can bring better performance. However, the teacher is the computational bottleneck in the whole structure, and the inference time of the teacher model increases significantly as the masking ratio decreases. So it is unreasonable to reduce the masking ratio hardly. By default, we choose 75\% as the masking ratio of the teacher model.
\par{\noindent \textbf{Distill manner.}}
As feature alignment can be divided into direct alignment and generation alignment, they can be used individually or in combination way. We compared the parallel combination strategy with the serial combination strategy in Table~\ref{tab:method}, the difference between the two ways is whether the mapping functions are shared or not. We consider that the serial alignment outperforms others as it makes it less difficult for the generation alignment.
\par{\noindent \textbf{Loss function.}}
For the loss function applied for feature alignment, we compared L1 loss and MSE loss. The performance of the former was 71.93\% and that of the latter was 72.38\%. Therefore, the MSE loss is our default setting.
\subsection{Main Results and Analysis}
\begin{table}[t!]
    \tablestyle{3pt}{1.1}
    \centering
    \begin{tabular}{y{60}x{54}x{48}x{48}}
    \toprule
    \textbf{Method} &\textbf{200 epochs} & \textbf{400 epochs} & \textbf{800 epochs} \\
    \midrule
    VideoMAE-B~\cite{videomae}& 66.4 & 67.9 & 69.6 \\
    \textbf{AMD-B~(ours)} & \textbf{72.4} & \textbf{72.8} & \textbf{73.3}\\
    \bottomrule
    \end{tabular}
    \vspace{-0.5em}
    \caption{\textbf{The effect of training schedule on SSV2.}}
    \label{tab:train_epochs}
    \vspace{-1em}
\end{table}
\begin{table}[t!]
\tablestyle{3pt}{1.1}
\begin{tabular}{lccc}
\toprule
\textbf{Method} &   \textbf{K400} $\rightarrow$ \textbf{SSV2} & \textbf{K400} $\rightarrow$ \textbf{UCF} & \textbf{K400} $\rightarrow$ \textbf{HMDB}  \\
\midrule
VideoMAE-B~\cite{videomae}   & 68.5 & 96.1  & 73.3 \\
$\textbf{AMD-B~(ours)}$ & $\textbf{72.6}~(\uparrow 4.1)$ & $\textbf{97.1}~(\uparrow 1.0)$ & $\textbf{79.6}~(\uparrow 6.3)$  \\
\bottomrule
\end{tabular}
\vspace{-0.5em}
\caption{\textbf{Comparison of the transfer performance.}}
\label{tab:transfer} 
\vspace{-1em}
\end{table}
\begin{table*}[t!]
\centering
\tablestyle{6pt}{1.0}
\begin{tabular}{c|l|c|c|c|c|c|c|c|c}

&\textbf{Method} & \textbf{Backbone} & \textbf{Extra data} & \textbf{Extra labels} & \textbf{Frames} & \textbf{GFLOPs}  & \textbf{Param} & \textbf{Top-1}  & \textbf{Top-5} \\
\shline\hline
\multirow{6}{*}{\rotatebox{90}{20-55M \# Param}}
&VideoMAE$_\text{2400e}$~\cite{videomae} & ViT-S & -- & \xmark & 16 & 57\x2\x3 & 22M & 66.8 & 90.3 \\
&MViTv1~\cite{mvit} & MViTv1-B & Kinetics-400 & \cmark & 64 & 455\x1\x3 & 37M & 67.7 & 90.9  \\
&TEINet$_{En}$~\cite{teinet} &  \footnotesize{ResNet50$_{\times 2}$} &  ImageNet-1K & \cmark & 8+16 &  99\x10\x3 & 50M & 66.5 & N/A \\
&TANet$_{En}$~\cite{tanet} &  \footnotesize{ResNet50$_{\times 2}$} & ImageNet-1K & \cmark & 8+16 & 99\x2\x3 & 51M & 66.0 & 90.1 \\
&SlowFast~\cite{slowfast} &  ResNet101 &  Kinetics-400 & \cmark & 8+32 & 106\x1\x3 & 53M & 63.1 & 87.6 \\
&\textbf{AMD}$_\text{800e}$~\textbf{(ours)} & ViT-S & -- & \xmark & 16 & 57\x2\x3 & 22M & \textbf{70.2} & \textbf{92.5} \\

\shline\hline
\multirow{7}{*}{\rotatebox{90}{55-100M \# Param}}
&VideoMAE$_\text{800e}$~\cite{videomae} & ViT-B & -- & \xmark & 16 & 180\x2\x3 & 87M & 69.6 & 92.0 \\
&VideoMAE$_\text{2400e}$~\cite{videomae} & ViT-B & -- & \xmark & 16 & 180\x2\x3 & 87M & 70.8 & 92.4 \\
&Video Swin~\cite{liu2021video}  & Swin-B & \footnotesize IN-21K+K400  & \cmark & 32 & 321\x1\x3 & 88M & 69.6 & 92.7  \\
&TDN$_{En}$~\cite{tdn} & \footnotesize{ResNet101$_{\times 2}$} & ImageNet-1K  & \cmark & 8+16 & 198\x1\x3 & 88M & 69.6 & 92.2   \\
&BEVT~\cite{bevt}  & Swin-B & \scriptsize{IN-1K+K400+DALLE}  & \xmark & 32 & 321\x1\x3 & 88M & 70.6 & N/A  \\
&DMAE$_\text{800e}^\dagger$~\cite{bai2022masked} & ViT-B & -- & \xmark & 16 & 180\x2\x3 & 87M & 70.0 & 92.5 \\
&\textbf{AMD}$_\text{800e}$~\textbf{(ours)} & ViT-B & -- & \xmark & 16 & 180\x2\x3 & 87M & \textbf{73.3} & \textbf{94.0} \\
\shline\hline
\multirow{7}{*}{\rotatebox{90}{>100M \# Param}}
&Motionformer~\cite{motionformer} & ViT-B &  \footnotesize IN-21K+K400  & \cmark & 16 & 370\x1\x3 & 109M  & 66.5 & 90.1 \\
&TimeSformer~\cite{timesformer} & ViT-B & ImageNet-21K & \cmark & 8 & 196\x1\x3 & 121M & 59.5 &  N/A \\
&ViViT FE~\cite{arnab2021vivit} & ViT-L & \footnotesize IN-21K+K400 & \cmark & 32 & 995\x 4\x3 & N/A & 65.9 & 89.9 \\
&VIMPAC~\cite{vimpac} & ViT-L & \scriptsize{HowTo100M+DALLE} & \xmark & 10 & N/A\x10\x3 & 307M & 68.1 & N/A \\
&Motionformer~\cite{motionformer} & ViT-L &  \footnotesize IN-21K+K400  & \cmark & 32 & 1185\x1\x3 & 382M  & 68.1 & 91.2 \\
&TimeSformer~\cite{timesformer} & ViT-L & ImageNet-21K & \cmark & 64  & 5549\x1\x3 & 430M & 62.4 & N/A \\
&\baseline{{VideoMAE$_\text{2400e}$~\cite{videomae}}} & \baseline{ViT-L} & \baseline{--} & \baseline{\xmark} & \baseline{16} & \baseline{597\x2\x3} & \baseline{305M} & \baseline{74.3} & \baseline{94.6} \\
\end{tabular}
\vspace{-1em}
\caption{\textbf{Comparison with previous works on SSV2.} Our AMD is pre-trained for 800 epochs on SSV2 using the serial feature alignment strategy. \xmark\ indicates no additional label information is used for pre-training. ''N/A'' means it is not available. ''$\dagger$'' denotes our implementation. DMAE is a distillation method. The teacher model is colored in \colorbox{baselinecolor}{gray}.}
\label{tab:ssv2}
\vspace{-1em}
\end{table*}
\begin{table*}[t!]
\centering
\tablestyle{6pt}{1.0}
\begin{tabular}{c|l|c|c|c|c|c|c|c|c}
&\textbf{Method} & \textbf{Backbone} & \textbf{Extra data} & \textbf{Extra labels} & \textbf{Frames} & \textbf{GFLOPs}  & \textbf{Param} & \textbf{Top-1}  & \textbf{Top-5} \\
\shline\hline
\multirow{7}{*}{\rotatebox{90}{20-55M \# Param}}
&VideoMAE~\cite{videomae} & ViT-S &-- & \xmark & 16 & 57\x5\x3 & 22M & 79.0 & 93.8 \\
&TSM~\cite{tsm} & ResNet50 &  ImageNet-1K  & \cmark & 128 & 65\x10\x3  & 24M  & 74.7 & 91.4    \\
&MViTv1~\cite{mvit} & MViTv1-S & -- & \xmark & 16 & 33\x5\x1 & 26M  & 76.0 & 92.1 \\
&ip-CSN~\cite{csn} & ResNet152 & -- & \xmark & 32 & 109$\times$10$\times$3 & 33M & 77.8 & 92.8 \\
&NL I3D~\cite{nonlocal} & ResNet50 &  ImageNet-1K  & \cmark & 128 & 282\x10\x3  & 35M  & 72.5 & 90.2    \\
&MViTv1~\cite{mvit} & MViTv1-B & -- & \xmark & 16 & 71\x5\x1 & 37M  & 78.4 & 93.5 \\
&\textbf{AMD}$_\text{800e}$~\textbf{(ours)} & ViT-S &-- & \xmark & 16 & 57\x5\x3 & 22M & \textbf{80.1} & \textbf{94.5} \\
\shline\hline
\multirow{9}{*}{\rotatebox{90}{55-100M \# Param}}
&TANet~\cite{tanet} &  ResNet152 & ImageNet-1K & \cmark &16 & 242\x4\x3 & 59M & 79.3 & 94.1 \\
&SlowFast~\cite{slowfast} &  R101+NL & --  & \xmark & 16+64 & 234\x10\x3 & 60M & 79.8 & 93.9 \\
&MAE-ST~\cite{feichtenhofer2022masked} & ViT-B & -- & \xmark & 16 & 180\x7\x3 & 87M & 81.3 & 94.9 \\
&VideoMAE$_\text{800e}$~\cite{videomae} & ViT-B & -- & \xmark & 16 & 180\x5\x3 & 87M & 80.0 & 94.4 \\
&VideoMAE$_\text{1600e}$~\cite{videomae} & ViT-B & -- & \xmark & 16 & 180\x5\x3 & 87M & 81.5 & 95.1 \\
&TDN$_{\text{En}}$~\cite{tdn} & ResNet101 & ImageNet-1K  & \cmark & 8+16 & 198\x10\x3 & 88M & 79.4 & 94.4   \\
&BEVT~\cite{bevt}  & Swin-B & IN-1K+DALLE  & \xmark & 32 & 282\x4\x3 & 88M & 80.6 & N/A  \\
&Video Swin~\cite{liu2021video}  & Swin-B & ImageNet-21K  & \cmark & 32 & 282\x4\x3 & 88M & 80.6 & 94.6  \\
&DMAE$_\text{800e}^\dagger$~\cite{bai2022masked} & ViT-B & -- & \xmark & 16 & 180\x5\x3 & 87M & 80.8 & 94.6 \\
&\textbf{AMD}$_\text{800e}$~\textbf{(ours)} & ViT-B & -- & \xmark & 16 & 180\x5\x3 & 87M & \textbf{82.2} & \textbf{95.3} \\
\shline\hline
\multirow{7}{*}{\rotatebox{90}{>100M \# Param}}
&Motionformer~\cite{motionformer} & ViT-B &  ImageNet-21K  & \cmark & 32 & 370\x10\x3 & 109M  & 79.7 & 94.2 \\
&TimeSformer~\cite{timesformer} & ViT-B & ImageNet-21K & \cmark & 96 & 590\x1\x3 & 121M & 78.0 & 93.7 \\
&ViViT FE~\cite{arnab2021vivit} & ViT-L & ImageNet-21K & \cmark & 128 & 3980\x 1\x3 & N/A & 81.7 & 93.8 \\
&VIMPAC~\cite{vimpac} & ViT-L & \scriptsize{HowTo100M+DALLE} & \xmark & 10 & N/A\x10\x3 & 307M & 77.4 & N/A \\
&Motionformer~\cite{motionformer} & ViT-L &  ImageNet-21K  & \cmark & 32 & 1185\x10\x3 & 382M  & 80.2 & 94.8 \\
& TimeSformer~\cite{timesformer} & ViT-L & ImageNet-21K  & \cmark & 96 & 8353\x1\x3 & 430M & 80.7 & 94.7 \\

&\baseline{VideoMAE$_\text{1600e}$}~\cite{videomae} & \baseline{ViT-L} & \baseline{--} & \baseline{\xmark} & \baseline{16} & \baseline{597\x5\x3} & \baseline{305M} & \baseline{85.2} & \baseline{96.8} \\

\end{tabular}
\vspace{-1em}
\caption{\textbf{Comparison with previous works on K400}. Our AMD is pre-trained for 800 epochs on K400 using the serial feature alignment strategy. \xmark\ indicates no additional label information is used for pre-training. ''N/A'' means it is not available. ''$\dagger$'' denotes our implementation. DMAE is a distillation method. The teacher model is colored in \colorbox{baselinecolor}{gray}.}
\vspace{-2em}
\label{tab:k400}
\end{table*}
\begin{table}[t!]
	\centering
	\tablestyle{5pt}{1.1}
    \begin{tabular}{l|c|c|c|c}
		\textbf{Method} &  \textbf{Backbone}  & \textbf{Extra labels} & \textbf{$T \times \tau$} & \textbf{mAP}   \\ 
		\shline
        supervised~\cite{slowfast} & SlowFast-R101 & \cmark & 8\x8 & 23.8 \\
        \hline
        CVRL~\cite{cvrl} & SlowOnly-R50   & \xmark & 32\x2  &  16.3 \\
        $\rho$BYOL$_{\rho=3}$~\cite{large} & SlowOnly-R50  & \xmark & 8\x8   & 23.4 \\
        $\rho$MoCo$_{\rho=3}$~\cite{large} & SlowOnly-R50  & \xmark & 8\x8   & 20.3 \\
        VideoMAE~\cite{videomae} & ViT-B  & \xmark & 16\x4  & 26.7 \\
        VideoMAE~\cite{videomae} & ViT-B  & \cmark & 16\x4  & 31.8 \\
        \textbf{AMD~(ours)} & ViT-B & \xmark & 16\x4  & \textbf{29.9} \\
        \textbf{AMD~(ours)} & ViT-B & \cmark & 16\x4  & \textbf{33.5} \\
	\end{tabular}
	\caption{\textbf{ Comparison with previous works on AVA v2.2.} ``\cmark'' means  we perform intermediate fine-tuning on K400 with \textit{labels} before transferred to AVA. $T \times \tau$ denotes the frame number and the sampling rate.
}\label{sota:ava}
\end{table}
\par{\noindent \textbf{AMD: small and strong MAE.}} 
The aim of our asymmetric distillation is to obtain a smaller yet stronger pre-trained MAE
model by feature distillation. As MAE benefits from more training epochs, we experimented AMD at different training epochs in Table~\ref{tab:train_epochs}. Compared with the VideoMAE base model, AMD performs better at a larger interval on different training schedules. It is significant to note that the performance of the teacher model on SSV2 is 74.3\%, and our best result at the training schedule of 800 epochs is only 1\% lower than it. Therefore, our AMD is a small and strong model thanks to the sufficient distillation.

\par{\noindent \textbf{Transfer learning: action recognition.}}
To verify the generalization ability of the distilled pre-trained model, we pre-trained AMD for 800 epochs on K400 and fine-tuned on SSV2, UCF101 and HMDB51. The result is presented in Table~\ref{tab:transfer}. Our asymmetric distillation approach significantly improves the generalisation ability of VideoMAE. It is notable that on the SSV2 dataset, the transfer performance is improved by 4.1\% compared to VideoMAE, which indicates that AMD reduces the risk of overfitting in transfer fine-tuning and demonstrates robust transfer performance.
\par{\noindent \textbf{Transfer learning: action detection.}}
Following the evaluation settings of VideoMAE, We also transfer the AMD pre-trained with the K400 to the action detection task on AVA in Table~\ref{sota:ava}. With unlabeled data, our AMD could achieve 29.9 mAP. If the intermediate fine-tuning is applied, our AMD could achieve 33.5 mAP. The results show that asymmetric distillation of VideoMAE can also further improve the downstream performance of the non-classification task, which further supports the robustness of our AMD.

\par{\noindent \textbf{Extreme case: the teacher model performs no masking.}} When we set the masking ratio of the teacher model to 0\%, we tried two settings, one where the student model does not perform any masking either, and the other where the student model remains at 90\% masking ratio. We perform the comparison with 200 epochs of training in Table~\ref{tab:no_masking}.

In the former case, we attempted to align the features directly without any masking of the teacher and the student. As the student is not masked, there is a retreat to the typical feature distillation. However, when the teacher's masking ratio was 45\%, AMD took significantly less time to achieve better performance~(72.5\% vs 72.4\%), which illustrates the efficiency of AMD. Since the teacher is the computational bottleneck of distillation, this direct feature alignment without mask imposes a higher computational overhead.

In the latter case, we attempted to adjust the masking ratio for the teacher to 0\% in AMD's default settings, which resulted in a relatively poor performance. We believe that the extreme setting would make the generation alignment difficult. We suggest choosing the teacher’s masking ratio from efficiency considerations as analyzed in \cref{abla}. And more analyses can be found in the appendix.
\begin{table}[t!]
    \tablestyle{4pt}{1.1}
    \centering
    \begin{tabular}{y{50}x{35}x{25}x{25}x{25}x{25}}
    \toprule
    \textbf{Method} &\textbf{Backbone}&\textbf{Tea.} & \textbf{Stu.} & \textbf{Top-1} & \textbf{Time}\\
    \midrule
    No-Masking &ViT-B& 0\% & 0\% & 72.4 & 32h \\
    AMD &ViT-B& 45\% & 90\% & \textbf{72.5} & \textbf{19h} \\
    AMD &ViT-B& 0\% & 90\% & 72.1 & 26h \\
    \bottomrule
    \end{tabular}
    \vspace{-0.5em}
    \caption{\textbf{Comparison with the teacher without masking.}}
    \label{tab:no_masking}
    \vspace{-1em}
\end{table}

\subsection{Application of AMD on Image Model}
We apply the asymmetric distillation to the image model ImageMAE~\cite{he2021masked} on ImageNet-1K~\cite{deng2009imagenet} to verify the generalizability of AMD. The masking ratio of the student and the teacher is 75\% and 50\% respectively. The official ViT-B and ViT-L pre-trained models are adopted for the teacher model. The results are presented in Table~\ref{tab:image_AMD}. when we adapt MAE-B as the teacher, AMD can achieve a comparable performance of 82.1\% based on ViT-S. when we adapt MAE-L as the teacher, AMD can achieve a classification accuracy of 84.6\% based on ViT-B, surpassing ImageMAE by 1.0\%.

\subsection{Comparison with the Symmetric Method}
To reveal the effect of context information in distillation, we compared our AMD with the symmetric method DMAE~\cite{bai2022masked}.
In the image domain, in Table~\ref{tab:image_AMD}, AMD outperforms DMAE by 2.8\% with MAE-B as the teacher model. And it remains better than 0.6\% with MAE-L as the teacher model.
In the video domain, in Table~\ref{tab:ssv2}, with the same teacher model and training length, AMD outperformed DMAE by a margin of 3.3\% in the SSV2 dataset. in Table~\ref{tab:k400}, AMD still outperforms DMAE by 1.4\% in the K400 dataset. Due to the asymmetric masking, the teacher model could capture more context information, which is beneficial for distillation and the serial alignment exploits context information better. However, the symmetric masking structure merely exploits the stronger feature representation capability of the teacher model. More analyses can be found in the appendix.
\begin{table}[t!]
	\centering
	\tablestyle{7pt}{1.1}
    \begin{tabular}{y{64}|x{35}|x{35}|x{40}}
		\textbf{Method} &  \textbf{Student}  & \textbf{Teacher} & \textbf{Top-1 Acc}\\
		\shline
        ImageMAE~\cite{he2021masked} & ViT-B  & -  & 83.6 \\
        ImageMAE~\cite{he2021masked} & ViT-L  & -  & 85.9 \\
        \hline
        SSTA~\cite{wu2022self} & DeiT-S & DeiT-B & 81.4 \\
        DMAE~\cite{bai2022masked} & ViT-S & MAE-B & 79.3 \\
        G2SD w/o S.D~\cite{huang2023generic} & ViT-S & MAE-B & 82.0 \\
        \textbf{AMD~(ours)} & ViT-S & MAE-B & \textbf{82.1} \\
        \hline
        DMAE~\cite{bai2022masked} & ViT-B & MAE-L & 84.0 \\
        \textbf{AMD~(ours)} & ViT-B & MAE-L & \textbf{84.6} \\
	\end{tabular}
        \vspace{-0.5em}
	\caption{\textbf{Performance of AMD applied to image model.}}
    \label{tab:image_AMD}
\vspace{-1em}
\end{table}
\subsection{Comparison with the State of the Art}
We compare the previous state-of-the-art results with our AMD on K400 and SSV2 in Table~\ref{tab:ssv2} and Table~\ref{tab:k400} respectively. We pre-trained AMD for 800 epochs for comparison based on both 16-frame vanilla ViT-S and ViT-B. We divided the models into three groups by using 55M and 100M as the boundaries for the number of model parameters. The results show that in the first two groups, AMD achieves the best results with a relatively small number of parameters. 

In the group below 55M, our AMD achieves the top-1 accuracy of 70.2\% on SSV2 and 80.1\% on K400 with no extra data used. Based on ViT-S, our AMD with 800 epochs of pre-training outperforms the VideoMAE with 2400 epochs of pre-training by 3.4\% on SSV2 and by 1.1\% on K400.

In the group above 55M, our AMD achieves the top-1 accuracy of 73.3\% on SSV2 and 82.2\% on K400 without extra data used for pre-training. Based on ViT-B, our AMD with 800 epochs of pre-training outperforms the VideoMAE with 800 epochs of pre-training by 3.7\% on SSV2 and by 2.2\% on K400. It is worth noting that our AMD surpasses some ViT-L based methods~\cite{timesformer,arnab2021vivit,motionformer,vimpac}.

\section{Conclusion}
In this paper, we have proposed an asymmetric masked distillation framework, termed as AMD, for pre-training relatively small foundation models with autoencoding. The asymmetric masking strategy allows the teacher model to capture more context information to transfer to the student model. We then proposed a customized feature alignment distillation method to take the advantage of the asymmetry, which can better exploit context information. 
Our AMD can yield smaller foundation models with excellent generalisation capabilities. We apply AMD to both ImageMAE and VideoMAE to demonstrate its effectiveness and versatility, obtaining impressive results in image classification and action recognition with a ViT-B backbone.

\paragraph {\bf Acknowledgements.} {This work is supported by the National Key R$\&$D Program of China (No. 2022ZD0160900), the National Natural Science Foundation of China (No. 62076119, No. 61921006),  and Collaborative Innovation Center of Novel Software Technology and Industrialization.}

\appendix
\section{Appendix}
In this supplementary material, we provide more details of our AMD and present more experiment results. Specifically, we provide more implementation details in our experiments in Section \ref{sec:impl}. Then, we present the differences between the different alignment methods in Section \ref{sec:arche}. After this, We discussed on the reconstruction task in Section \ref{sec:recon}. Then, we continued our analysis of the teacher performing no masking in Section \ref{sec:no_mask}. Next, we discussed the masking of the student model in Section \ref{sec:stu_mask}. Finally, we continued our analysis of the comparison with DMAE in Section \ref{sec:compare}.

\section{Implementation Details}\label{sec:impl}
We conduct the experiments with 32 A100-80G GPUs for pre-training on SSV2 and K400. Additionally, we fine-tune the SSV2 with 16 GPUs, the K400 and the AVA with 32 GPUs. All ablation experiments conduct with 16 GPUs. The experiments on UCF101 and HMDB51 both worked with 8 GPUs. Our implementation is based on VideoMAE~\cite{videomae} and follows the data augmentation settings of pre-training and fine-tuning.  To speed up model training and improve the stability, we perform the repeated sampling~\cite{Hoffer2020AugmentYB}. The training schedule we give is the total number of times a sample has been sampled. The pre-training settings on the SSV2 and K400 datasets are showm in Table \ref{tab:pretrain}.

\begin{table}[ht]
\tablestyle{5pt}{1.1}
\small
\begin{tabular}{y{85}|x{60}x{60}}
\textbf{config} & \textbf{SSV2} & \textbf{K400} \\
\shline
optimizer & \multicolumn{2}{c}{AdamW} \\ 
learning rate & \multicolumn{2}{c}{1.2e-3}\\
weight decay & \multicolumn{2}{c}{0.05} \\
optimizer momentum & \multicolumn{2}{c}{$\beta_1, \beta_2{=}0.9, 0.95$~\cite{Chen2020c}} \\
batch size & \multicolumn{2}{c}{2048} \\
repeated sampling~\cite{Hoffer2020AugmentYB} & \multicolumn{2}{c}{4} \\
learning rate schedule & \multicolumn{2}{c}{cosine decay~\cite{coslr}} \\
epochs & \multicolumn{2}{c}{800} \\
warmup epochs & \multicolumn{2}{c}{40} \\
sampling rate & 2 & 4 \\
flip augmentation & \multicolumn{2}{c}{\emph{no}} \\
augmentation & \multicolumn{2}{c}{MultiScaleCrop~\cite{tsn_journal}} \\
\end{tabular}
\caption{AMD pre-training setting for both ViT-S and ViT-B backbone.}
\label{tab:pretrain} 
\vspace{-0em}
\end{table}

\begin{figure*}[ht]
  \centering
   \includegraphics[width=0.95\linewidth]{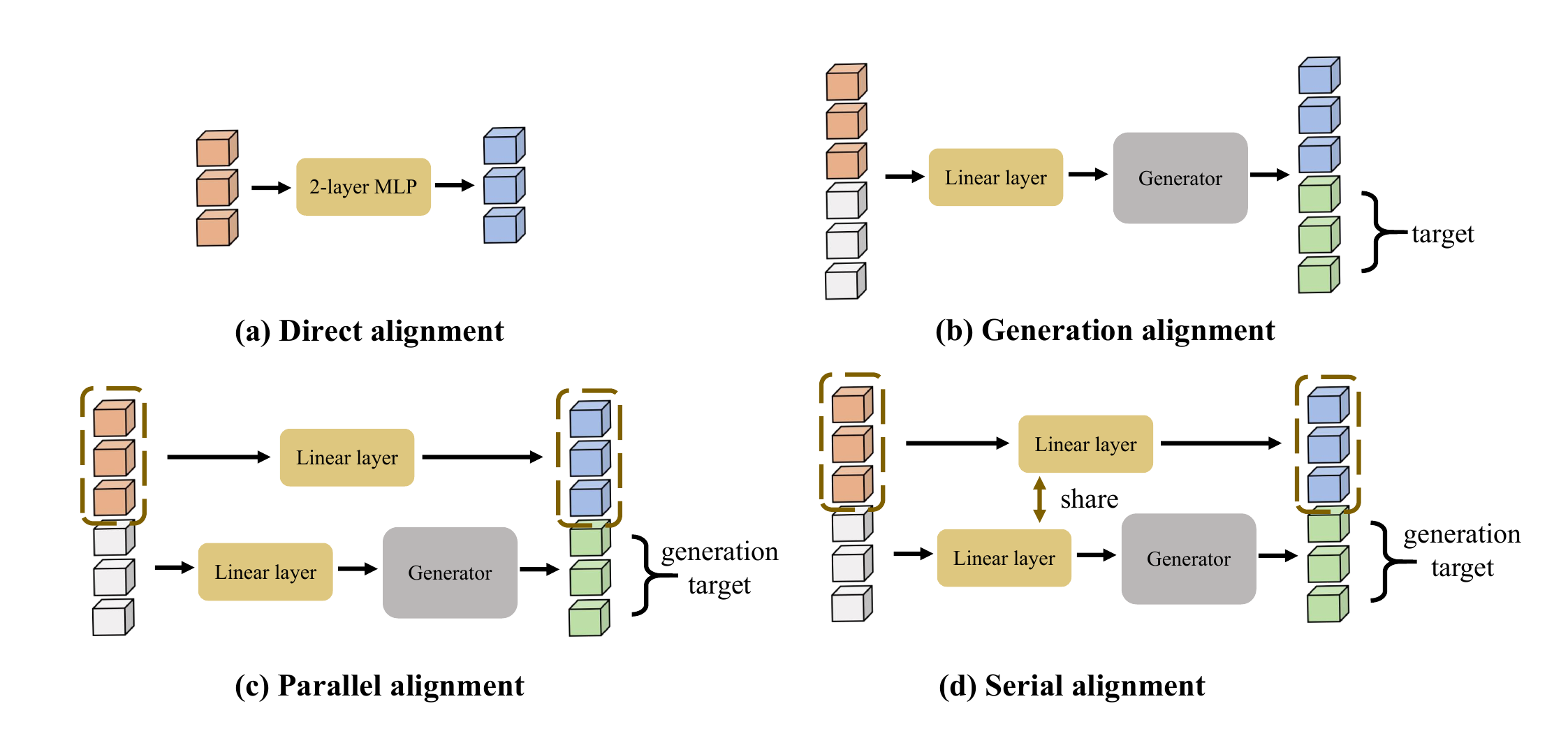}
   \vspace{-1em}
   \caption{We apply four feature alignment methods in our work, with the serial alignment being our default setting.}
   \vspace{-1em}
   \label{fig:4_arch}
\end{figure*}

\begin{table}[t!]
\tablestyle{5pt}{1.1}
\small
\begin{tabular}{y{85}|x{60}x{60}}
\textbf{config} & \textbf{SSV2} &\textbf{K400} \\
\shline
optimizer & \multicolumn{2}{c}{AdamW} \\
base learning rate & 1e-3~(S),~5e-4(B) & 1e-3~(S),7e-4~(B) \\
weight decay & \multicolumn{2}{c}{0.05} \\
optimizer momentum & \multicolumn{2}{c}{$\beta_1, \beta_2{=}0.9, 0.999$} \\
batch size & 512~(S,B) & 512~(S),1024~(B)\\
learning rate schedule & \multicolumn{2}{c}{cosine decay~\cite{coslr}} \\
warmup epochs & \multicolumn{2}{c}{5} \\
training epochs &  40~(S), 30~(B) & 150~(S), 90~(B) \\
sampling rate & \textit{sparse}~\cite{tsn_journal} & 4 \\
repeated sampling~\cite{Hoffer2020AugmentYB} & \multicolumn{2}{c}{2} \\
flip augmentation & \emph{no} & \emph{yes} \\
RandAug~\cite{randaugment}  & \multicolumn{2}{c}{(9, 0.5)} \\
label smoothing~\cite{label_smoothing}  & \multicolumn{2}{c}{0.1} \\
mixup~\cite{mixup}  & \multicolumn{2}{c}{0.8}  \\
cutmix~\cite{cutmix}  & \multicolumn{2}{c}{1.0} \\
drop path & \multicolumn{2}{c}{0.1} \\
head dropout  & \multicolumn{2}{c}{\textit{None}} \\
layer-wise lr decay~\cite{beit}  & 0.7~(S),0.75~(B) & 0.75~(S,B)  \\
\end{tabular}
\caption{AMD fine-tuning setting of SSV2 and K400.}
\vspace{-2em}
\label{tab:s_fineutne_ss_k400}
\end{table}

\begin{table}[t!]
\tablestyle{1pt}{1.0}
\small
\begin{tabular}{y{85}|x{48}x{48}x{45}}
\textbf{config} & \textbf{HMDB51} &\textbf{UCF101} & \textbf{AVA} \\
\shline
optimizer & \multicolumn{3}{c}{AdamW} \\
base learning rate & 1e-3 & 5e-4 & 2.5e-4\\
weight decay & \multicolumn{3}{c}{0.05} \\
optimizer momentum & \multicolumn{3}{c}{$\beta_1, \beta_2{=}0.9, 0.999$} \\
batch size & 128 & 256 &128\\
learning rate schedule & \multicolumn{3}{c}{cosine decay~\cite{coslr}} \\
warmup epochs & \multicolumn{3}{c}{5} \\
training epochs &  60 & 100 &30 \\
sampling rate & 2 & 4 & 4 \\
repeated sampling~\cite{Hoffer2020AugmentYB} & 2 & 2 & \textit{no} \\
flip augmentation & \multicolumn{3}{c}{\emph{yes}} \\
RandAug~\cite{randaugment}  &  (9, 0.5) & (9, 0.5) & -- \\
label smoothing~\cite{label_smoothing}  & 0.1 & 0.1 & -- \\
mixup~\cite{mixup}  & 0.8 & 0.8 & --  \\
cutmix~\cite{cutmix}  & 1.0 & 1.0 & -- \\
drop path & 0.1 & 0.2 & 0.2\\
head dropout  & 0.5 & 0.5 & \textit{None} \\
layer-wise lr decay~\cite{beit}  &0.75 &0.70 &0.75  \\
\end{tabular}
\vspace{-.2em}
\caption{Fine-tuning setting of HMDB51, UCF101 and AVA.}
\label{tab:s_fineutne_3dataset}
\vspace{-2em}
\end{table}

\par{\noindent \textbf{SSV2.}}
We pretrain AMD on SSV2 for 800 epochs by default. For fine-tuning, we perform the spare sampling~\cite{tsn_journal} and report the $2$ clips \x $3$ crops evaluation results and the settings are shown in Table \ref{tab:s_fineutne_ss_k400}.
\par{\noindent \textbf{K400.}}
 We pretrain AMD on K400 for 800 epochs by default. For fine-tuning, we report the $5$ clips \x $3$ crops evaluation results and the settings are shown in Table \ref{tab:s_fineutne_ss_k400}.
 \par{\noindent \textbf{ HMDB51 and UCF101.}}
We only fine-tune the model pre-trained on K400 to the HMDB51 and UCF101 dataset. We report the $5$ clips \x $3$ crops evaluation results and the settings are shown in Table \ref{tab:s_fineutne_3dataset}.
\par{\noindent \textbf{AVA.}}
We refer to the most classic two-stage structure to detect key frames of the video. In the first stage, we use the box detected in AIA~\cite{tang2020asynchronous}. While in the second stage, we use the ViT backbone to classify the objects detected in the first stage. Follwing VideoMAE, the short side size of the input is resized to 256 pixels. The ground-truth person boxes are only used for training. In term of the inference, we use the detected boxes with confidence $\geq0.8$. The settings are shown in Table \ref{tab:s_fineutne_3dataset}.
\section{Different alignment methods Details}\label{sec:arche}
Four alignment strategies are described in this paper, with specific structural details described below and the structures are shown in Figure \ref{fig:4_arch}.
\par{\noindent \textbf{Direct alignment.}}
We employ a 2layer MLP for alignment, where the hidden layer has the same dimension as the teacher model and the activation function is GELU~\cite{Hendrycks2016GaussianEL}. Additionally, the features to be aligned have not been normalised.
\par{\noindent \textbf{Generation alignment.}}
We applied a decoder-like generator to align teacher features, where the number of \texttt{[MASK]} tokens is the number of tokens that the teacher has more than the student. A linear projection layer is needed to align the teacher's dimension before the student features are fed into the generator. And the features used to calculate the alignment loss is also those features that the teacher have more of than the student.
\par{\noindent \textbf{Parallel alignment.}}
We have combined the two alignment methods in a parallel way, where the direct alignment part uses only a simple linear projection layer. It is worth noting that the projection layer of the two alignment methods do not share parameters.
\par{\noindent \textbf{Serial alignment.}}
We combine the two alignment methods in a serial way as our default setting, and the two aligned linear projection layers share parameters, which can reduce the difficulty of generation alignment.
\begin{table}[t]
    \tablestyle{1pt}{1.0}
    \centering
    \begin{tabular}{y{40}x{40}x{40}x{60}x{40}}
    \toprule
    \textbf{Method} & \textbf{Model} &\textbf{Epochs}&\textbf{Reconstruction} & \textbf{Top-1} \\
    \midrule
    AMD & ViT-B & 800 & \xmark & 73.0 \\
    AMD & ViT-B & 800 & \cmark & \textbf{73.3} \\
    \bottomrule
    \end{tabular}
    \caption{We compared the results with and without the reconstruction task with 800 epochs of training.}
    \label{tab:no-recon}
    \vspace{-2em}
\end{table}
\begin{figure*}[t]
  \centering
   \includegraphics[width=1.0\linewidth]{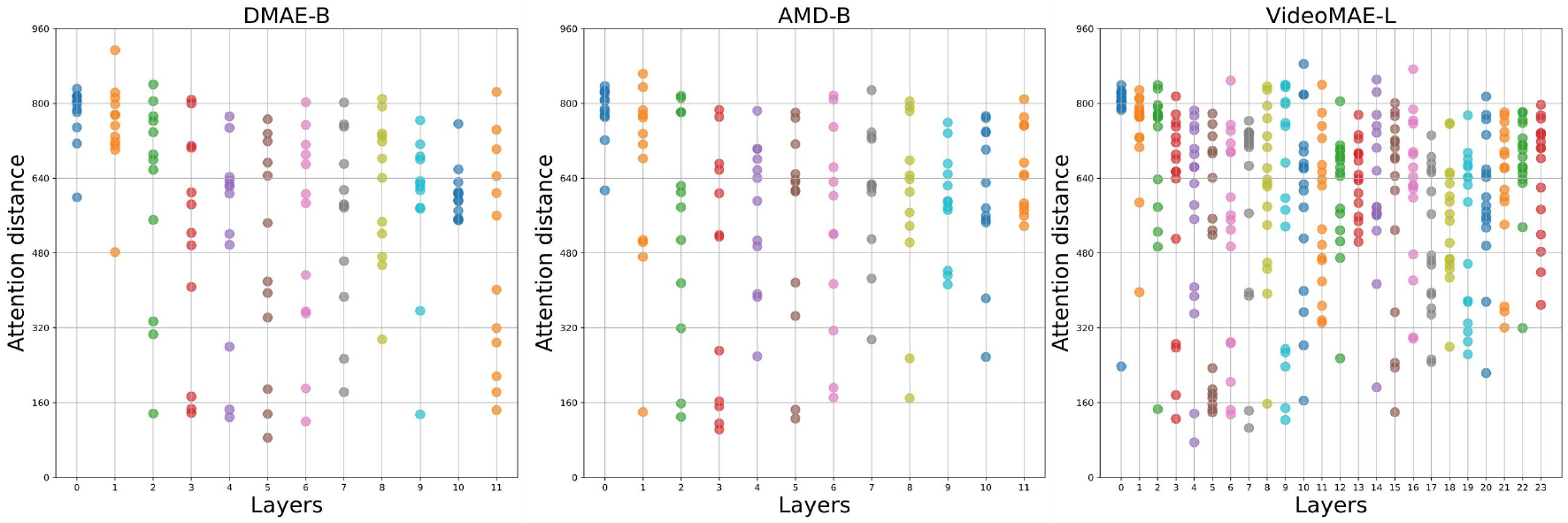}
   \vspace{-0em}
   \caption{\textbf{The average attention distance in different attention heads at each layer depth.} Distances are calculated over 16 frames, and frame spacing is calculated over the maximum distance of each frame. Results are averaged over SSV2 test set.}
   \label{fig:attn_dis}
   \vspace{-0em}
\end{figure*}

\section{Discussion on reconstruction task}\label{sec:recon}
To verify the effect of the reconstruction task in distillation, we have made a comparison in Table \ref{tab:no-recon}. The results show that distillation using the reconstruction task performs better. We consider that the reconstruction task provides a regularisation for model distillation and allows students to learn more semantic information that is beneficial for generalisation, which also allows AMD to benefit from a longer training schedule.

\section{Analysis of the masking of the student}\label{sec:stu_mask}
Note that the VideoMAE's optimal masking ratio on the reconstruction task is 90\%, and AMD also focuses on the pre-training, so we fixed the masking ratio of the student model at 90\% in order not to damage the reconstruction difficulty. We supplement an experiment with a student masking ratio of 80\% and a teacher masking ratio of 75\%, whose accuracy is 73.1\% after 800 epochs of training, lower than the default setting~(73.3\%).
\section{Analysis of teacher performing no masking}\label{sec:no_mask}
We found that the performance degradation occurs when the teacher masking is extremely low. We think that there might be a conflict in our training goals. The conflict becomes more apparent under an extremely low masking ratio of the teacher model. Our training aims to do two things: 1) reconstructing the image pixels and 2) aligning with the teacher's features. However, a low masking ratio in the teacher model means it covers more global information. This can lead to a mismatch with the student model's reconstruction task. We have noticed that the reconstruction loss rises when the masking ratio of the teacher model becomes quite low, which may support our conjecture about the conflict.

Furthermore, we supplemented a experiment with a teacher's masking ratio of 25\%, which resulted in 72.3\%. So the peak in accuracy might occur roughly at a teacher masking ratio of 45\%.
However, when the teacher's masking ratio was reduced from 60\% to 45\%, its training cost increases but its gain is very limited.
So we suggest choosing the teacher’s masking ratio from efficiency considerations.
\begin{figure}[t]
  \centering
   \includegraphics[width=1.0\linewidth]{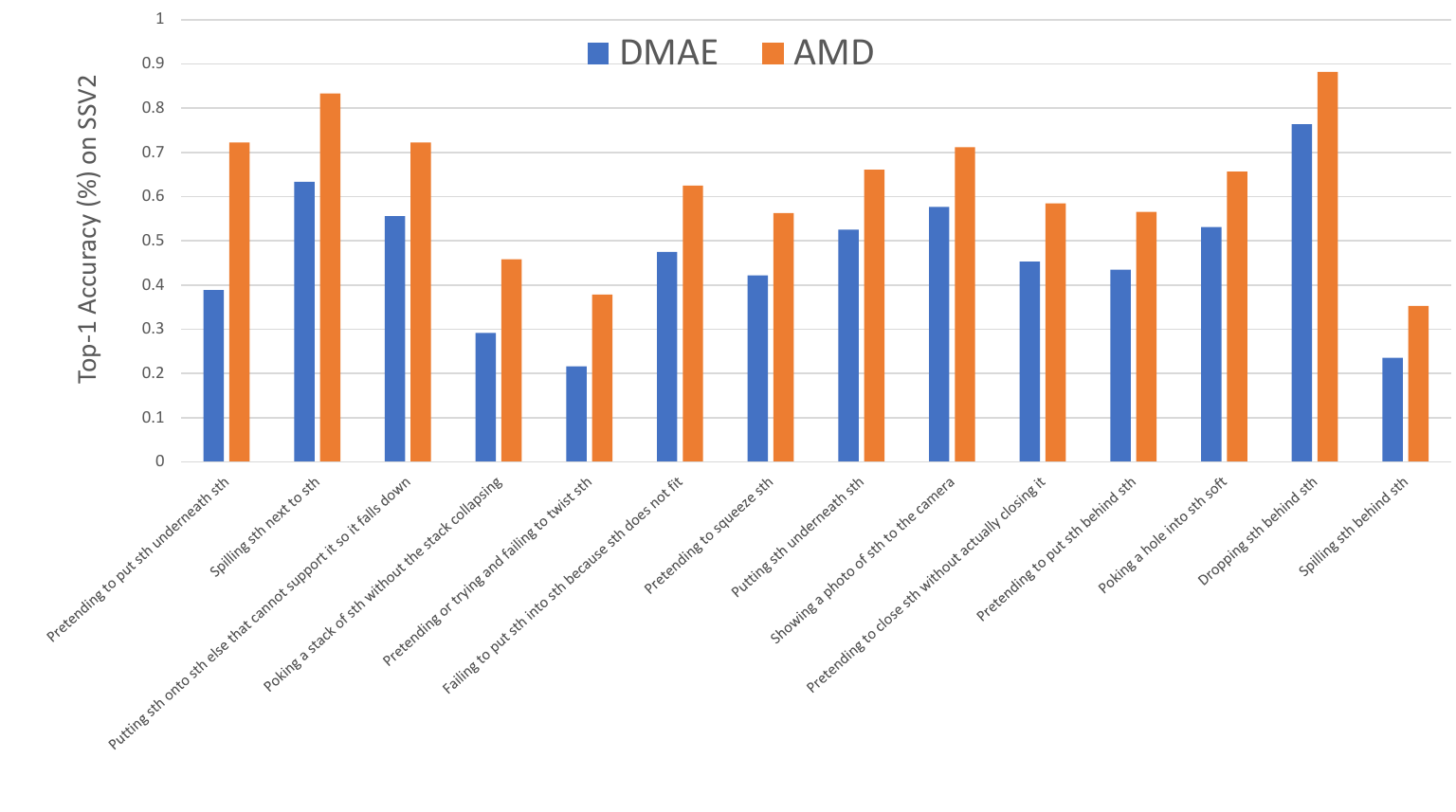}
   \vspace{-2em}
   \caption{\textbf{Detailed breakdown of accuracy comparison between AMD and DMAE by categories.} We checked the performance gap on SSV2 in terms of categories on the test set.}
   \label{fig:breakdown}
\end{figure}
\section{Comparison with DMAE}\label{sec:compare}
Overall, the two main differences between AMD and DMAE are asymmetric masking and generation alignment which are discussed in Table 1. In addition, to understand the distinct impacts of the AMD and DMAE masking 
distillation strategies on the video model pre-training, we provided a comparison of 14 categories in SSV2 with the most accuracy difference in Figure \ref{fig:breakdown}. It shows that AMD has a stronger ability to infer object interactions, spatial relationships, and action outcomes.

Furthermore, we examined the average attention distance of DMAE, AMD and VideoMAE~(the teacher model) to reveal the properties of models in Figure \ref{fig:attn_dis}. We find that at shallow layers, each model has diverse attention heads which means model's attention is both local and global. While at deep layers especially the last layer, most attention heads of AMD and VideoMAE tend to extract global informations, which is different from DMAE. In the video domain, the more global attention means that the model is able to capture more global information about the action, which is beneficial for action recognition. Therefore AMD is better at tasks that require temporal understanding, which is due to the asymmetric masking strategy that allows the teacher model to see more contextual information.

{\small
\bibliographystyle{ieee_fullname}
\bibliography{egbib}
}
\end{document}